# Perspectives and Ethics of the Autonomous Artificial Thinking Systems

Joël Colloc

Normandie Univ, UNIHAVRE, CNRS, IDEES, 76600  Le Havre France
joel.colloc@univ-lehavre.fr,
WWW home page: http://www.cirtai.org/spip.php7rubrique389

**Abstract:** The feasibility of autonomous artificial thinking systems needs to compare the way the human beings acquire their information and develops the thought with the current capacities of the autonomous information systems. Our model uses four hierarchies: the hierarchy of information systems, the cognitive hierarchy, the linguistic hierarchy and the digital informative hierarchy that combines artificial intelligence, the power of computers models, methods and tools to develop autonomous information systems. The question of the capability of autonomous system to provide a form of artificial thought arises with the ethical consequences on the social life and the perspective of transhumanism.

## 1. Introduction

The recent progress of the cognitive sciences, neurosciences, computing and robotics boosts the project of autonomous artificial being capable of thinking. The feasibility of autonomous artificial thinking systems needs to compare the way the human beings acquire their information and develops the thought with the current capacities of the autonomous information systems. Our model is based on four hierarchies: the hierarchy of information systems, stemming from the systemic theory, supplies indicators of complexity and autonomy; the cognitive hierarchy describes the sub-symbolic acquisition and the emergence of our personal experience, our knowledge, whereas the linguistic hierarchy builds the speech describing the knowledge acquired in terms of concrete and abstract objects on the environment and on oneself. The digital informative hierarchy relies on the necessary concepts, models, methods and tools to build autonomous information systems. At last, we discuss the nature of artificial thinking.

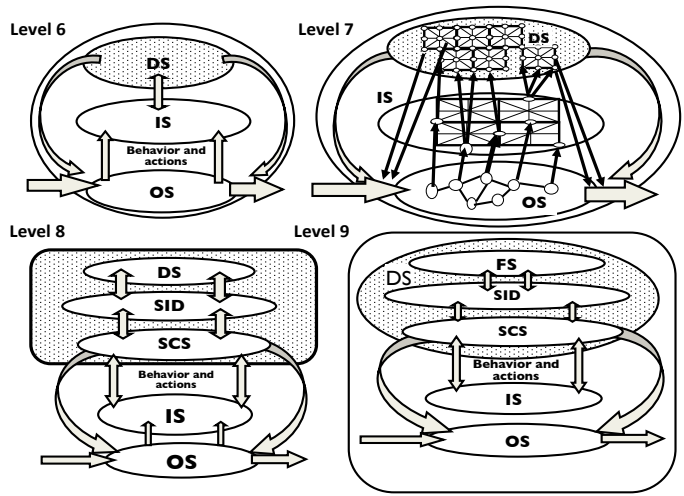

**Caption:**
Level 6: Operating, Information Decision system
Level 7: The system coordonates its actions of different agents
Level 8: The system imagines and elaborates new means of decision
Level 9: The system finalizes
**Acronyms:**
**SCS:** Selection and Coordination System
**DS:** Decision System
**FS:** Finalization System
**IS:** Information System
**SID:** System of Imagination and Design
**OS:** Operating System

Figure 1: The levels 6 to 9 of the model of complex system (J-L. Le Moigne, 1990)

## 2. The hierarchy of information systems

Jean-Louis Le Moigne describes the evolution of information systems. Like epigenesis, the following level adds features to the previous one. By conciseness, the levels 6 to 9 are presented figure 1. On level 6, the system becomes is able to memorize his decision O.I.D (Operating system (O), Information system (I), System of Decision (D)). On level 7, the system coordinates numerous decisions of actions at all the time t, concerning its internal activity and regulation and the external information from and to the environment. On level 8, the system is endowed with a sub-system of imagination and design (SIC). On level 9, the system is able to decide on its decision and to determine the positive and negative aspects of its actions. This finalization of a complex system is closed to the human thought (SF) which confers him an autonomy of decision allowing him to set his own goals. They correspond to the autonomous multi-agents systems.

## 3. Integration of the sensory ways and motricity: emergence of objects

The neurophysiology of the sensory ways is described with details in (Arbib and Hanson 1988),(Gazzaniga et al., 2001) and (Purves et al., 1997).For all the sensory ways (vision, hearing, smell, taste, touch, ...) we find globally an ascending similar organization: peripheral sensors, transduction in electric signal, the transmission (+/- long) by nerves in the brainstem or by pairs of cranial nerves (V, VII, IX, X), the relay in a specific thalamic area, a projection in a specialized cortical zone and associative areas common to several sensory modalities. In every level, numerous recurring neurons constitute the feedback loops of the cerebral cortex that are essential to do the predictions of the future expected perceptions as coherent with those recently memorized at each level of details: from short-lived elements of perception called "percepts" up to durably memorized objects (figure 2 and 3). The motricity works similarly by downward ways and is strongly entangled with the sensory ways and also have feedback loops. The cerebral cortex functioning is rather uniform on its whole surface (Mountcastle, 1978). Seeing an apple actives the smell of the fruit, its shape, consistency (in connection with the motricity) and the word « apple » to speak about it. Smelling (blindly) the odor of apple activates a stereotypical image of apple which we would expect to see. All these predictions is possible because we memorized a global unified perception of the object apple bound with the other various sensory, linguistic modalities which correspond to it (Hawkins and Blakeslee, 2005), (Minsky, 1988).

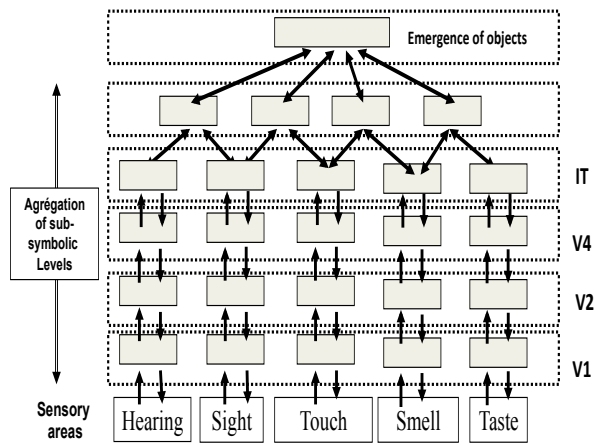
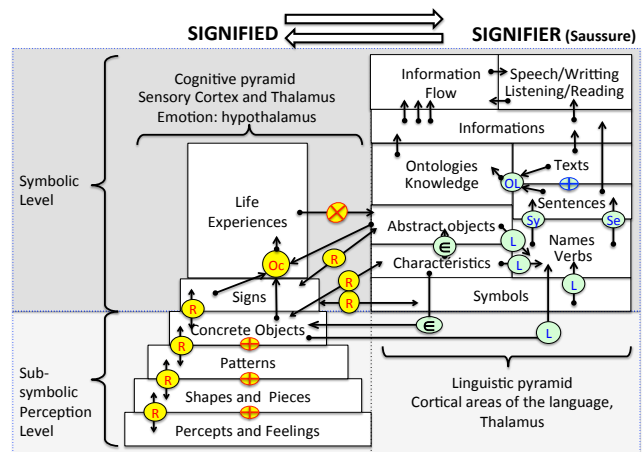

Figure 2: Sensory emergence of known objects (Hawkins and Blakeslee, 2005)

Figure 3: Cognitive / linguistic hierarchy (Colloc, 2016)

The figure 2 shows the various cortical layers V1, V2, V4 that integrate the sensory modalities and associative zones in the inféro-temporal cortex (IT) and that are activated when an apple is recognized in any manner. This activation is then transmitted in a downward way in the other sensory or linguistic areas which are going to predict patterns corresponding to levels of details more and more fine to confirm or counter that they are corresponding in each of the other senses. We can so know if we are feeling a real apple or only a partial representation as in the famous Magritte's painting entitled « *Ceci n'est pas une pomme* » (this is not an apple).To this anatomo-physiological organization corresponds a hierarchy of the sensory information passed on and integrated into the cortical level figure 3. The hippocampus is the zone situated at the top of the mnesic pyramid. This cerebral structure is at the origin of the matching and the interconnection of the various sensory modalities and the memorization of the characteristics of the new unknown objects (Hawkins and Blakeslee, 2005). The figure 3 describes an abstract model of the human information system (Colloc, 2016). In his left part we find the sensory and cognitive pyramid leading to the emergence of concrete objects and to the memorization of knowledge and experiences on the objects of the world. At every level recognition operators (R) confirm or counter the coherence in the flow of the perceptions. Percepts is primary elements perceived by the organs of the senses. For the vision, it is points, of directed segments (vertical, oblique, horizontal ...) which establish in the field of vision of the forms elementary as those described in the theory of "*geons*" proposed by Hummel and Biederman in 1987 (Biederman, 1987),(Boucart, 1996) and the hypothalamic emotional aspects concerning them (Damasio, 1995). The neurohormones, secreted mainly by the hypothalamus and the amygdalae, influence the interpretation of the sensory information. The hypothalamus regulates the set of the balances of the body (hunger and thirst/satisfaction, pleasure/pain, ...) (Vincent, 1996),(Vincent, 2002). The amygdalae intervene in the circuits of the reward and the motivation. For Alain Prochiantz, there is no organ of the thought, " there n is no thought without body, nor of body without thought" (Prochiantz, 2001), he so joins Francisco Varela (Varela et al., 1993). The cognitive operators ⊕ composition and (R) Recognition establish signs which base the experience of patterns and behavior of the concrete objects at successive levels of compositions (part left of the figure 3). The operator of synthesis ⊗ Predicted the characteristics or the behavior of the similar objects and exploit the following operators: the operator of generalization (G) groups together similar objects in a category. The operator of deduction (Dd) predicts the characteristics and the behavior of objects to be looked for from the shapes and already observed patterns. The induction operator (In) search for predictable properties in unknown objects because of their resemblance with already known objects. The operator of abduction (Ab) detects missing or abnormal properties in usually known objects. The operator of subsumption (Sb) connects an object type more specialized in a more general already known object type. The analogy operator detects similarity between the structures, the characteristics or the behavior of known and memorized objects.

**The linguistic hierarchy:** The concrete objects do not need to be put into words to be recognized, the language appears later with the maturation of the nervous system. Jean Piaget underlines that the language is gradually set up from the 12th to the 18th months of the life of the child (Piaget, 1992). The sign language being older than the spoken language, it is earlier set up in the encephalon and precedes the language in the verbal flow (Jacquet-Andrieu and Colloc, 2014). The perception, the emergence, the recognition of the concrete objects and the thought are previous to the implementation of the language, they are thus considered as sub-symbolic, they result from the memorization of the experiences of perception. The Haeckel's theory :"The ontogeny recapitulates the philogenesis" finds a new lighting by the discovery of the genes of the development (Hox) (Prochiantz, 2001). In particular, the development of the sensory ways takes place during the embryogenesis and continues after the birth in the presence of the stimulations of the environment. The absence of visual stimulation of kittens in the birth prevents the maturation of the vision and causes a partial or total agenesis of the visual ways (Shatz, 1992). Also, the other intellectual functions would be organized according to an intellectual protomap (similar to that proposed by Brodman) which determines the migration of neurons and the maturation of the sensory and motor ways. The linguistic pyramid (part right of figure 3) establishes the symbols, the characteristics, the names of objects abstracted from the concrete objects or the other pre-existent abstract objects. It determines semantics, the syntax of the sentences allowing to establish input flows (reading / listening) and output (speaking / writing) of texts and flows of information (verbal and nonverbal). The linguistic operators (Ol) are: the lexical operator (L) in charge of naming and associating a word with the concrete and abstract objects and with their characteristics or behavior. The operator of membership ($\in$) (has a) establishes that a characteristic or a behavior belongs to a concrete or abstract object. The semantic operator defines himself the meaning of a word (name, verb, adverb) that is with which concrete or abstract object is the meant according to (de Saussure, 1916). The syntactic operator (Sy) determines in a given language the possible syntaxes of the sentences. The objects of the world are meaningful by their name, the actions by verbs and their characteristics by adjectives, the quite ordered by the syntax rules of a grammar and operators like generalization (is-a), causality (caused by), composition (is-part-of), temporality (succeeds, precedes)... They allow to create new abstract objects memorized in their turn and then combined with others in an incremental way. Our model is always incomplete and reducing but it shows the interactions between the perception, the cognition, the language and the feelings. It tries to reconcile the structuralist approach of the linguistics initialized by Ferdinand de Saussure with the current neurophysiological and neuro-psychological approaches (de Saussure, 1916).

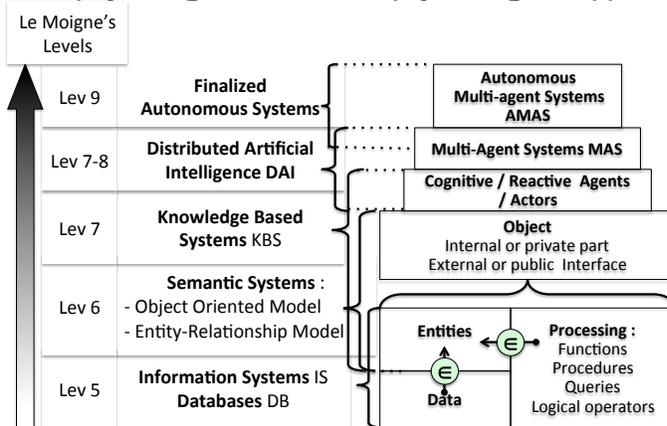

**Caption**
**5.** Objects encapsulate ($\in$) data or attributes, and procedures to make necessary arithmetics or logic calculations and deductions from the data values.
**6.** Semantic systems implements entities and then objects
**7-8.** Distributed Knowledge bases implement cognitive agents and ontologies cooperation.
**9.** Conscious autonomous multi-agent systems (CAMAS) has an internal memory to store facts of consciousness. It is able to choose its objectives and then schedule its goals and actions.

Figure 4 : Digital information hierarchy (Colloc, 2016)

## 5 The digital information hierarchy

Information relies on the triptych: data, information and knowledge (Abiteboul, 2012).
Many knowledge bases were developed like MYCIN (Buchanan and Shortliffe, 1984) and SIAMED (Colloc, 1985). Logic is used to classify objects as initially shown by (Carroll, 1896) and languages like LISP (McCarthy,1960) or PROLOG (Colmerauer, 1996). Semantic models (object oriented and agents) allows to build knowledge and ontologies more efficiently (Shen et

al.,2015). The semantic Web allow to access a considerable quantity of knowledge and data on the world and the people (level 6-7). Multi-agent systems (MAS) are used to build simulations: A community of numerous agents ants solve problems of search for paths in complex graphs thanks to their stigmergic behavior (level 7). MAS allows to to coordinate mutiple ontologies and knowledge agents to support clinical decision in medicine (Shen et al., 2015). Alain Cardon proposes a psychic system that relies on knowledge in psychology and capable of generating flows of thoughts that take place in the temporality under the shape of organized groups of processes to build the artificial psychic system and its interactions (Cardon, 2011). This conscious autonomous multi-agent system (CAMAS) is able to choose its objectives and set its goals to achieve them (level 9).

**6. Conclusion Can computers think ?**

Computers could think in a very different way than human, as a plane flights differently but faster than a bird (Pitrat, 1990). The Zen is a total holism, the world cannot be absolutely divided into parts. The dilemma is that for every object of the world, according to master Zen Mummon: "we cannot express him with words and we cannot express him without the words". According to the Buddhism : to trust the words to reach the truth is equivalent of trusting an always incomplete formal system (Hofstadter, 1985). For Jiddu Khrishnamurti, our consciousness is common to all the humanity: All the human beings think that contribute to build it. He considers the individualism, the ego as an obstacle to understand the consciousness with rare moments of clarity ("*insight*"). "*The thought is a movement in the time and the space. The thought is memory, memory of the past things. The thought is the activity of the knowledge, the knowledge which was gathered through millions of years and stored in the form of memory in the brain.*" (Krishnamurty, 2005). There are two forms of thinking: -The first is a reaction of the memory which contains the knowledge, the result of the experience from the beginning of humanity (phylogenesis) and since our birth (epigenesis) in a loop: experience, knowledge, memory, thought, action, experience... necessarily limited by the time, it is used every day, rational, individualistic, power-hungry and of progress submissive to the knowledge which accumulates, in the words which divide and this division is responsible for all the suffering, for all the troubles of the world. -The alternative thinking: the action-perception where in rare occasions, we pay simply our attention to the world, without interpreting it, without naming anything, virgin of any prejudice, knowledge and especially spontaneously, by living this moment without thinking of it and without the will. For example: the direct perception of a wonderful landscape of mountain one morning with all our senses uses our complete attention where we forget ourselves and banish the use of the words (Krishnamurty, 2005). Such a full perception requests only the left lower part: the sub-symbolic cognitive pyramid (figure 3) and not the linguistic pyramid. The first way of thinking is a fatal vicious circle implemented soon in computers that will think better than human beings! And there nothing prevents that a computer invents a new religion at the origin of new sufferings for the humanity. The language during the evolution would have made lose to the man his spontaneity in the immediate perception of the world such as it is. The digital technology, which strengthens the symbolic nature of our relation to the world, is doubtless going to establish the peak of our ignorance. The realization of CAMAS able to think is now possible. These systems have access to Internet and will quickly become more powerful than the human beings with disturbing consequences for the future of the humanity. In conclusion, we have shown that computers can think in the first way only. The main characteristic of humanity is to still be able to think also in the alternative way but how long before losing our essential faculty ?